\documentclass[letterpaper]{article} % DO NOT CHANGE THIS
\usepackage{aaai24}  % DO NOT CHANGE THIS
\usepackage{times}  % DO NOT CHANGE THIS
\usepackage{helvet}  % DO NOT CHANGE THIS
\usepackage{courier}  % DO NOT CHANGE THIS
\usepackage[hyphens]{url}  % DO NOT CHANGE THIS
\usepackage{graphicx} % DO NOT CHANGE THIS
\usepackage{natbib}  % DO NOT CHANGE THIS AND DO NOT ADD ANY OPTIONS TO IT
\usepackage{caption} % DO NOT CHANGE THIS AND DO NOT ADD ANY OPTIONS TO IT
\usepackage{algorithm}
\usepackage{algorithmic}
\usepackage{tabularx}
\usepackage{amssymb}
\usepackage{amsmath}
\usepackage{bm}
\usepackage{gensymb}

\usepackage{newfloat}
\usepackage{listings}
\DeclareCaptionStyle{ruled}{labelfont=normalfont,labelsep=colon,strut=off} % DO NOT CHANGE THIS
\floatstyle{ruled}
\newfloat{listing}{tb}{lst}{}
\floatname{listing}{Listing}
\title{Monocular 3D Hand Mesh Recovery via Dual Noise Estimation}
\author{
    Hanhui Li\textsuperscript{\rm 1}, Xiaojian Lin\textsuperscript{\rm 1}, Xuan Huang\textsuperscript{\rm 1}, Zejun Yang\textsuperscript{\rm 2}, Zhisheng Wang\textsuperscript{\rm 2}, Xiaodan Liang\textsuperscript{\rm 1}\thanks{Corresponding author: xdliang328@gmail.com.}
}
\affiliations{
    \textsuperscript{\rm 1}Shenzhen Campus of Sun Yat-sen University, Shenzhen, China\\
    \textsuperscript{\rm 2}Tencent, Shenzhen, China\\
}

\begin{document}

\maketitle

\begin{abstract}
Current parametric models have made notable progress in 3D hand pose and shape estimation. However, due to the fixed hand topology and complex hand poses, current models are hard to generate meshes that are aligned with the image well. To tackle this issue, we introduce a dual noise estimation method in this paper. Given a single-view image as input, we first adopt a baseline parametric regressor to obtain the coarse hand meshes. We assume the mesh vertices and their image-plane projections are noisy, and can be associated in a unified probabilistic model. We then learn the distributions of noise to refine mesh vertices and their projections. The refined vertices are further utilized to refine camera parameters in a closed-form manner. Consequently, our method obtains well-aligned and high-quality 3D hand meshes. Extensive experiments on the large-scale Interhand2.6M dataset demonstrate that the proposed method not only improves the performance of its baseline by more than 10$\%$ but also achieves state-of-the-art performance. Project page: \url{https://github.com/hanhuili/DNE4Hand}.
\end{abstract}

\section{Introduction}
Recent advances in parametric human models \cite{pavlakos2019expressive} have facilitated human-centric applications, such as artificial intelligence generated content, human avatars, and virtual talking heads. With parametric models like \cite{romero2017embodied}, reconstructing 3D hand meshes from images becomes plausible and convenient. This attracts considerable attention and extensive research has been conducted to improve the accuracy and speed of the parametric model fitting process \cite{zhang2021interacting,yu2023acr,meng20223d,moon2023bringing,li2022cliff,chen2022mobrecon}.

Nevertheless, reconstructing well-aligned hand meshes from single-view images is still challenging because of the following two reasons: (i) Challenging factors like depth ambiguity, self/inter-hand occlusions, and complicated hand motions hinder estimation accuracy. (ii) Even worse, the pre-defined hand topology in parametric models further restricts hand mesh deformations, and consequently the parametric models are not flexible enough to represent various hands.

Non-parametric hand models \cite{lin2021end,lin2021mesh,lin2022mpt,jiang2023a2j} seem to be a possible solution for the above issues. With the powerful representation learning ability \cite{tian2023recovering}, current non-parametric methods can predict mesh vertices directly. This provides great flexibility and in practice methods of this type usually yield better performance, compared with parametric models. However, without leveraging the hand structural prior, these methods are prone to producing severe artifacts and broken meshes. 

It is natural to consider combining parametric models and non-parametric models to leverage the structural advantages of the former and the flexibility of the latter. Methods belonging to this paradigm \cite{tang2021towards,li2022interacting,ren2023two,yu2023overcoming,moon2023bringing} have been proposed recently. However, as far as we are concerned, most current methods seek a deterministic manner, such as predicting the deviations of vertices and parameters \cite{tang2021towards}. This makes them hard to explore the solution space thoroughly. Note that recovering 3D hand meshes from monocular images is an ill-posed problem, which means multiple meshes can be associated with the same 2D observation. Therefore, a deterministic model may be ineffective to tackle this task.

To tackle the above issues, we propose to tackle monocular hand mesh recovery in a probabilistic framework. Particularly, given a monocular input image, we adopt an off-the-shelf parametric model as the baseline to obtain the coarse hand meshes. We then refine the coarse hand meshes by jointly estimating the noise of vertices and their image-plane projections, since they are highly related. We design a progressive framework to do so, in which image-aligned features are leveraged to estimate the parameters governing the noise distributions. With the estimated distributions, we adopt the reparameterization trick to generate multiple samples and estimate their confidence. In this way, we can leverage the sample with the highest confidence to optimize the hand vertices and their projections. Moreover, given the refined vertices and their 2D coordinates, we also propose a closed-form solution to refine camera parameters. Consequently, our method can generate hand meshes that are aligned with images well. Our experiments on the Interhand2.6M dataset show that the proposed method can boost the quality of the coarse meshes significantly, and achieve the state-of-the-art performance.

Our contributions can be summarized as follows:

\noindent$\bullet$ To the best of our knowledge, this paper proposes the first probabilistic 2D and 3D noise estimation framework for the monocular hand mesh recovery task.

\noindent$\bullet$ An effective network architecture is introduced to realize the dual noise estimation process. This network leverages image-aligned features and multiple samples to enhance the coarse meshes generated by baseline parametric models.

\noindent$\bullet$ The proposed method is validated on the large-scale Interhand2.6M dataset and outperforms conventional methods. 

\section{Related Work}

\subsection{Parametric Hand Models}

Parametric models for 3D hand meshes have gained considerable attention because it provides the convenient structural/geometric prior of hands. Extensive methods have been proposed for fitting parametric models, such as attention modules \cite{zhang2021interacting,yu2023acr}, inverse kinematic solvers \cite{shetty2023pliks,li2023hybrik}, hand disentanglement \cite{meng20223d,moon2023bringing}, and 2D-3D projection \cite{li2022cliff}. The comprehensive review of parametric models can refer to \cite{tian2023recovering}. Parametric models can be roughly divided into regression based methods and optimization based methods. Regression based methods estimate the parameters directly while optimization based methods usually involve an online optimization process. Parametric models are restricted by their pre-defined hand templates and are inflexible to model hands of various poses and geometry.

\subsection{Non-parametric Hand Models}
Early non-parametric models aim at predicting hand joints from depth maps and point clouds \cite{cheng2022efficient,deng2022recurrent}. With the recent advances in network architectures, non-parametric models that predict 3D hand vertices become popular. For instance, transformers and graph neural networks \cite{lin2021end,lin2021mesh,lin2022mpt,jiang2023a2j} have been proposed for mesh reconstruction. Since non-parametric models do not rely on the fixed hand topology, they are more flexible and easier to be aligned with images. However, without the structural prior of hands, non-parametric models also suffer from distorted and spiky reconstruction results. 
 
\subsection{Hybrid Hand Models}
It is reasonable to construct hybrid models and leverage the advantages of both parametric and non-parametric models. Several pioneering studies have been conducted to achieve this goal. For example, IntagHand \cite{li2022interacting} utilizes the topology of MANO to construct the graph representation of vertices. It also defines graph attention modules to model vertex dependencies. \citealt{ren2023two} incorporate the MANO model into a point cloud network for pose estimation. \citealt{yu2023overcoming} propose to estimate joints first via non-parametric models and then infer MANO parameters based on joints. \citealt{moon2023bringing} introduce a network to predict relative translation between two MANO hands. 

The proposed method differs from traditional methods because of its probabilistic unified modeling of vertices and their 2D coordinates. With the proposed method, we can achieve the mutual and progressive refinement between vertices and 2D coordinates.

\subsection{Implicit Hand Models}

Except for explicit representations, recent studies also try to explore implicit functions (e.g., signed distance function, \citealt{park2019deepsdf}) to represent 3D hands. A notable advantage of implicit functions is that they are continuous and disentangled from spatial resolutions. This advantage indicates that implicit functions can generalize to arbitrary hands. Several implicit hand models have been proposed, such as LISA \cite{corona2022lisa}, AlignSDF \cite{chen2022alignsdf}, Im2Hands \cite{lee2023im2hands}, HandNeRF \cite{guo2023handnerf}, and Hand Avatar \cite{chen2023hand}. However, compared with explicit models, the computational cost of implicit models is more expensive. 

\section{Methodology}
\begin{figure*}[t]
  \centering
  \includegraphics[width=\textwidth]{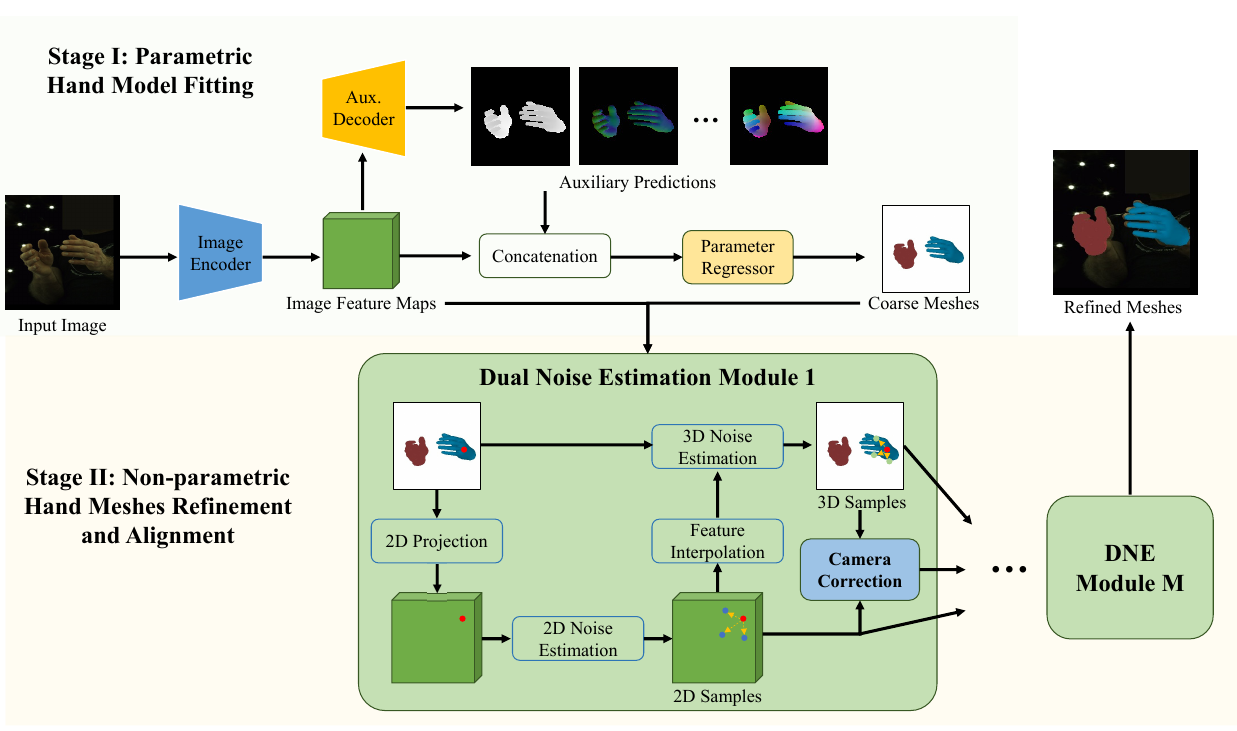}
  \caption{The overall framework of the proposed method. It is a two-stage that first generate coarse meshes and then refine them via multiple dual noise estimation modules.}
  \label{fig:frame}
\end{figure*}

\subsection{Overall Architecture}
The architecture of our method is shown in Figure \ref{fig:frame}. It consists of a coarse mesh fitting stage and a refinement stage. Particularly, given a single-view image as input, we adopt ResNet-50 \cite{he2016deep} as the image encoder to extract the feature maps $\mathbf{F}$ of size $H \times W \times C$. To better leverage the geometric and semantic information in the image, we adopt a 2D convolutional block to predict five auxiliary maps, including the depth map $\mathbf{a}_{1} \in \mathbb{R}^{H \times W}$, the normal map $\mathbf{a}_{2} \in \mathbb{R}^{H \times W \times 3}$, the joint heat map $\mathbf{a}_{3} \in \mathbb{R}^{H \times W \times 42}$ (each hand has 21 joints), the DensePose map \cite{guler2018densepose} $\mathbf{a}_{4} \in \mathbb{R}^{H \times W \times 3}$, and the part semantic map $\mathbf{a}_{5} \in \mathbb{R}^{H \times W \times 34}$ (16 parts for each hand, plus one background class). We merge $\mathbf{F}$ and these five auxiliary maps via channel-wise concatenation followed by another 2D convolutional block. For conciseness, we still denote the merged feature maps as $\mathbf{F}$. 

$\mathbf{F}$ are then fed into a baseline fitting model to predict the parameters of MANO, including the pose coefficient $\bm{\theta} \in \mathbb{R}^{16\times 6}$ \cite{zhou2019continuity}, the shape coefficient $\bm{\beta} \in \mathbb{R}^{10}$, and the intrinsic camera parameters $\mathbf{c} \in \mathbb{R}^{2\times 2}$ for each hand. Inspired by \citealt{yu2023acr}, we utilize 2D convolutions to predict parameter maps and accumulate the parameters via spatial softmax. With the fitted parameters, we generate the initial coarse hand meshes.

To refine the coarse meshes and better align them with images, we introduce the dual noise estimation (DNE) module. The core of DNE is to conduct mesh refinement and alignment jointly in a denoising process. To this end, the DNE module first refines the image-plane projections of coarse vertices. Then the DNE module obtains the image-aligned features via interpolation and uses them to estimate 3D vertex deviations. Furthermore, based on the correspondences between 3D vertices and their 2D projections, the DNE module also adopts a closed-form solution to refine the camera parameters. The detailed architecture of the DNE module is presented in the next section. The above refinement process is conducted progressively via multiple DNE modules and in practice we find that more DNE modules yield more significant performance gains. 

\subsection{Dual Noise Estimation}

\textbf{Formulation}. Given an arbitrary vertex ${\bf{v}} \in \mathbb{R}^3$ and its corresponding 2D coordinate ${\bf{u}} \in \mathbb{R}^2$, our proposed DNE module can be formulated as follows:
% \begin{equation}
%     {\bf{s}}({\bf{v}}_{xy} + {\bm{\varepsilon} _{3d}}) + {\bf{t}} = {{\bf{u}}} + {\bm{\varepsilon} _{2d}},
% \label{eq:projection}
% \end{equation}
\begin{equation}
    \Pi ({\bf{v}} + {\bm{\varepsilon} _{3d}}, {\bf{c}}) = {{\bf{u}}} + {\bm{\varepsilon} _{2d}},
\label{eq:dual_noise}
\end{equation}
where $\Pi$ denotes the 2D projection of $\bf{v}$ given the intrinsic camera parameters ${\bf{c}}$. Note that $\bf{u}$ is not necessarily obtained by $\Pi$. As we demonstrate later, $\bf{u}$ can be regressed from the image feature maps directly. ${\bm{\varepsilon} _{3d}}$ and ${\bm{\varepsilon} _{2d}}$ are the 3D and 2D noise terms that need to be estimated. To ensure our network is differentiable, we assume both ${\bm{\varepsilon} _{3d}}$ and ${\bm{\varepsilon} _{2d}}$ follow a certain distribution, on which we can apply the reparameterization trick \cite{kingma2013auto}. In this paper, we adopt the Gaussian distribution to model ${\bm{\varepsilon} _{3d}}$ and ${\bm{\varepsilon} _{2d}}$, namely,
\begin{equation}
    \begin{split}
        {\bm{\varepsilon}_{3d}} \sim \mathcal{N}(\bm{\mu}_{3d}, \gamma|\bm{\mu}_{3d}| + \delta) \\
        {\bm{\varepsilon}_{2d}} \sim \mathcal{N}(\bm{\mu}_{2d}, \gamma|\bm{\mu}_{2d}| + \delta)
    \end{split}
\label{eq:gaussian}
\end{equation}
where $\gamma, \delta > 0$ are hyperparameters that control the scale and margin of noise, respectively. Based on Eq. (\ref{eq:gaussian}), we can sample multiple ${\bm{\varepsilon} _{3d}}$ and ${\bm{\varepsilon} _{2d}}$ to better explore the solution space during training, and set ${\bm{\varepsilon} _{3d}} = \bm{\mu}_{3d}$ and ${\bm{\varepsilon} _{2d}} = \bm{\mu}_{2d}$ for inference. This makes the proposed method differ from traditional methods that only estimate 2D/3D deviations. Our task now turns to estimating appropriate $\bm{\mu}_{3d}$ and $\bm{\mu}_{2d}$.

%Moreover, we assume all the DNE modules form a Markov chain so that we can consider the noise estimation in each DNE module separately.

$\bm{\mu}_{2d}$ \textbf{Estimation}. Image-aligned features that are obtained via feature interpolation are leveraged to estimate $\bm{\mu}_{2d}$. Particularly, we consider two types of 2D coordinates in the interpolation process, including (i) the 2D projections of vertices (i.e., $\Pi ({\bf{v}}, {\bf{c}})$) and (ii) those that are regressed directly from the image feature maps. The intuition behind such a combination is that features obtained by the first type can maintain the hand structure and be robust to outliers, while those of the second type are more flexible. 

To regress 2D coordinates from the image feature maps $\mathbf{F}$, we reshape $\mathbf{F}$ to $(HW)\times C$ and adopt two consecutive multilayer perceptrons (MLPs) to transform $\mathbf{F}$ to $N \times C$ first and then $N \times 3$, where $N=778$ is the number of vertices of one MANO hand. Let ${\mathbf{f}}_p$ and ${\mathbf{f}}_r$ denote the $C$-dimensional interpolated feature vector with the projected and regressed coordinates, respectively. We consider the following transformation $\phi: \mathbb{R}^{2C} \to \mathbb{R}^{2}$ to obtain the per-vertex mean of 2D noise:
\begin{equation}
    \bm{\mu}_{2d} = \phi({\mathbf{f}}_p, {\mathbf{f}}_r).
    \label{eq:mu2d}
\end{equation}
In our network, $\phi$ is implemented efficiently via feature concatenation followed by an MLP.

\begin{figure*}[t]
  \centering
  \includegraphics[width=\textwidth]{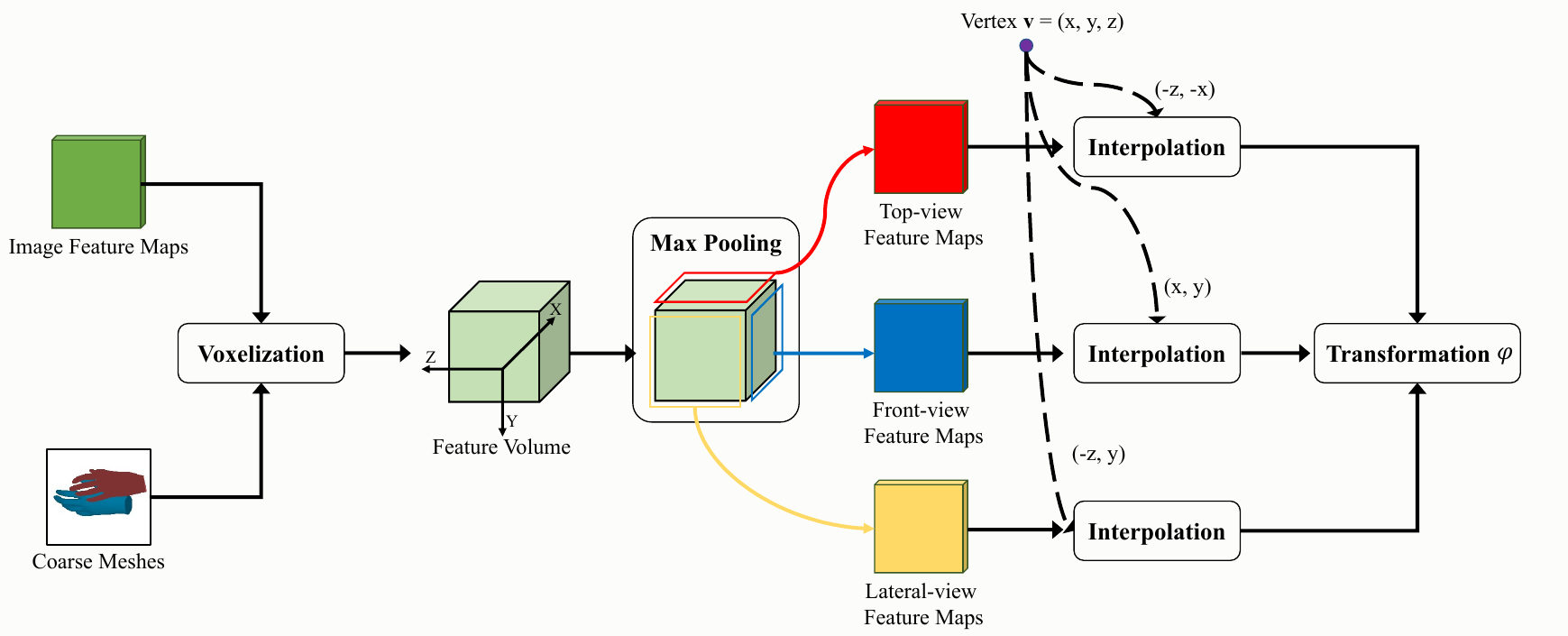}
  \caption{Architecture of the module for per-vertex mean of 3D noise estimation. It leverages three-view feature map disentanglement to alleviate depth ambiguity and occlusions.}
  \label{fig:3de}
\end{figure*}

$\bm{\mu}_{3d}$ \textbf{Estimation}. We also extract image-aligned features from $\mathbf{F}$ to estimate $\bm{\mu}_{3d}$. The updated 2D coordinate ${{\bf{u}}} + {\bm{\varepsilon} _{2d}}$ is used for feature interpolation. Considering that image-aligned features might be insufficient in tackling depth ambiguity and occlusions, here we propose a simple yet effective method to alleviate this problem. As shown in Figure \ref{fig:3de}, we first create a voxel grid from the hand meshes (with normalized 3D coordinates), so that features of spatially closed vertices can be aggregated into the same voxel. We then conduct max pooling along each of the three axes of the grid, to obtain the three-view (front, lateral, and top) projections of voxel features. 

The above multi-view feature projections help to achieve finer feature disentanglement compared with the single-view representation. Similar to Eq. (\ref{eq:mu2d}), the per-vertex mean of 3D noise can be estimated via a transformation $\varphi: \mathbb{R}^{3C} \to \mathbb{R}^{3}$ as follows:
\begin{equation}
        \bm{\mu}_{3d} = \varphi({\mathbf{f}}_{front}, {\mathbf{f}}_{lateral}, {\mathbf{f}}_{top}).
    \label{eq:mu3d}
\end{equation}
We also adopt an MLP to implement $\varphi$. Note that except for the MANO model, we do not impose any other constraint on the topology of vertices. This allows us to seamlessly adopt architectures that are more complicated than MLPs (e.g., Graph attentions, \citealt{li2022interacting}) to realize $\phi$ and $\varphi$.

\textbf{Camera Correction}. Last but not least, the updated $\bf{v}$ and $\bf{u}$ are used to refine the intrinsic camera parameters. We adopt the orthographic camera model and hence $\Pi ({\bf{v}}, {\bf{c}})$ can be defined as follows:
\begin{equation}
    \Pi ({\bf{v}}, {\bf{c}}) = {\bf{s}}{\bf{v}}(x,y) + {\bf{t}},
\label{eq:projection}
\end{equation}
where ${\bf{v}}(x,y)$ denotes the $x$ and $y$ coordinates of $\bf{v}$. ${\bf{s}}=(s_x, s_y)$ and ${\bf{t}} = (t_x, t_y)$ are the scaling factors and the principle point translations of the camera from the normalized device coordinate space to the image space\footnote{\url{https://pytorch3d.org/docs/cameras}}. Hence $\bf{c}$ can be represented as ${\bf{c}} = \left[ {\begin{array}{*{20}{c}}
{{s_x}},&{{t_x}}\\
{{s_y}},&{{t_y}}
\end{array}} \right]$ and the projection process in can be written as $u_x = {s_x}{v_x} + t_x$, $u_y = {s_y}{v_y} + t_y$. Substituting Eq. (\ref{eq:projection}) into Eq. (\ref{eq:dual_noise}), we estimate the intrinsic camera parameters by minimizing the following equation:
\begin{equation}
    \Sigma _{n = 1}^N||{{\bf{u}}_n} - {\bf{s}}{{\bf{v}}_n}(x,y) + {\bf{t}}||_2^2 + \xi (||{\bf{s}}||_2^2 + ||{\bf{t}}||_2^2),
    \label{eq:ridge_goal}
\end{equation}
where $\xi > 0$ is a hyperparameter for regularization. Here we omit ${\bm{\varepsilon} _{3d}}$ and ${\bm{\varepsilon} _{2d}}$ and use $\bf{v}$ and $\bf{u}$ to denote the updated vertex and its 2D coordinate. Eq. (\ref{eq:ridge_goal}) can be solved via ridge regression \cite{bishop2006pattern}, which has the following closed-form solution: 
\begin{equation}
    {\bf{c}'} = {({\bf{V}}_{}^{\top}{{\bf{V}}_{}} + \xi {\bf{I}})^{-1}}{\bf{V}}_{}^{\top}{\bf{U}_{}},
\end{equation}
where ${\bf{V}}$, ${\bf{U}} \in \mathbb{R}^{N \times 2}$ are the matrix representation of all vertices and their 2D coordinates, $\bf{I}$ is a $2 \times 2$ identity matrix.

\begin{figure*}[!t]
  \centering
  \includegraphics[width=\textwidth]{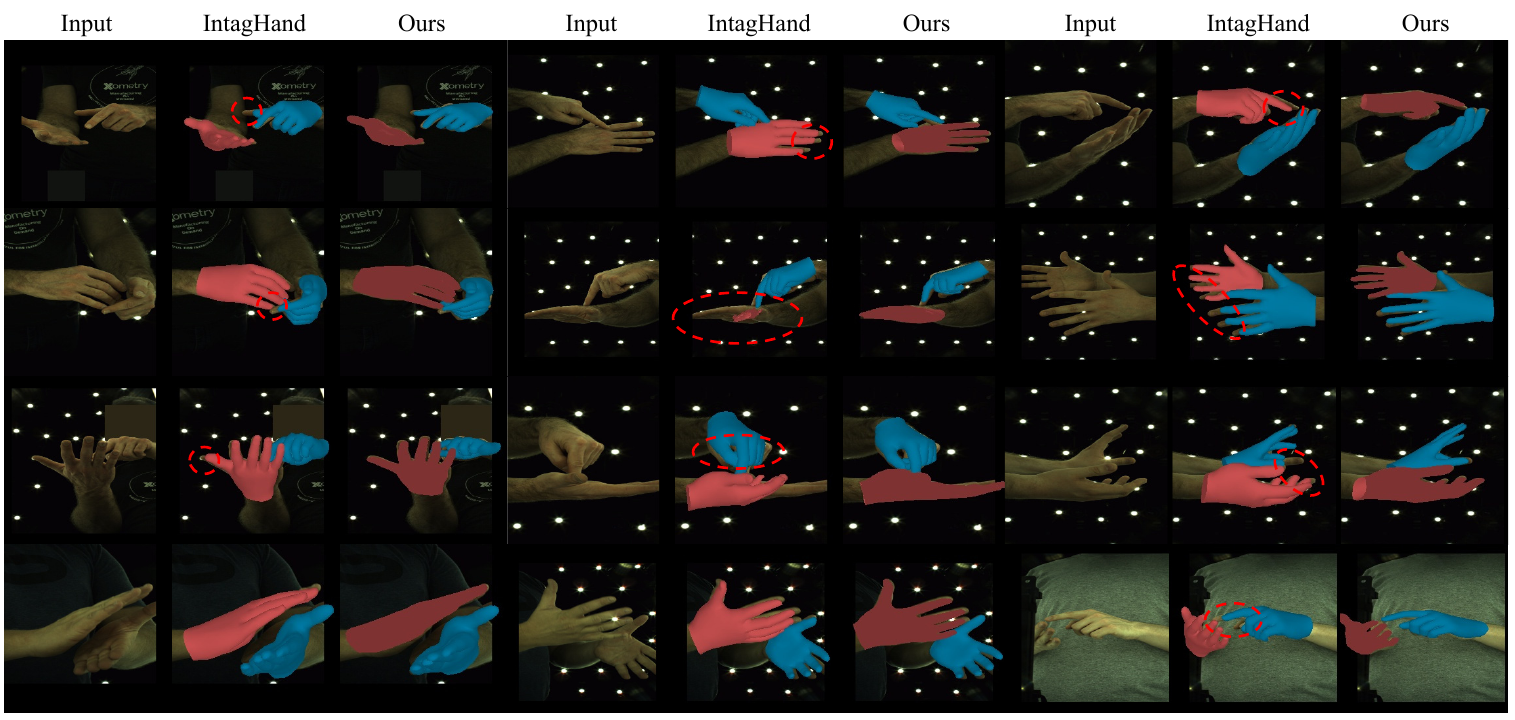}
  \caption{Visual comparison between the proposed method and the state-of-the-art non-parametric method (IntagHand). We highlight a few misaligned parts that are alleviated by the proposed method. Best viewed on screen.}
  \label{fig:qualcomp}
\end{figure*}

\subsection{Optimization}

The proposed network is fully differentiable and is trained via minimizing the following loss function:
\begin{equation}
    L = L_{aux} + L_{MANO} + L_{v},
    \label{eq:overall_optimization}
\end{equation}
where $L_{aux}$ denotes the loss for the five auxiliary tasks. ${L_{aux}}$ is formulated as follows:
\begin{equation}
    {L_{aux}} = \Sigma _{i = 1}^4 {\lambda}_{i}||{{\bf{a}}_i} - {\bf{a}}_i^g|{|_1} + {\lambda}_{5}CE({{\bf{a}}_5}, {\bf{a}}_5^g),
    \label{eq:loss_aux}
\end{equation}
where $||\cdot|{|_{1}}$ is the $l_1$ loss and $CE$ is the cross-entropy loss. Variables marked with the superscript $g$ are ground truths.
$L_{MANO}$ is the loss term defined on the shape and pose parameters of MANO:
\begin{equation}
    {L_{MANO}} = {\lambda}_{\beta}||{{\bm{\beta}}} - {\bm{\beta}}^g|{|_1} + {\lambda}_{\theta}||{{\bm{\theta}}} - {\bm{\theta}}^g|{|_1}.
    \label{eq:loss_mano}
\end{equation}
$L_v$ targets at mesh vertices and their 2D projections and is defined as follows:
\begin{equation}
\begin{split}
        {L_v} = \, & \lambda_{3D}\Sigma_{m = 1}^M \Sigma_{r = 1}^R {||{{\bf{v}}_{m,r}} - {{\bf{v}}^g}|{|_1}} \, + \\
    \, & \lambda_{2D}\Sigma_{m = 1}^M\Sigma_{r = 1}^R {||\Pi ({{\bf{v}}_{m,r}},{{\bf{c}}_m}) - \Pi ({{\bf{v}}_{m}},{{\bf{c}}^g})|{|_1}},
\end{split}
\label{eq:loss_v}
\end{equation}
where $M$ is the number of DNE modules and $R$ is the number of random samples in each DNE module. $\lambda _{1}, ..., \lambda _{5}, {\lambda}_{\beta}, {\lambda}_{\theta}, {\lambda}_{2D}, {\lambda}_{3D}$ are user-specified loss weights.

\section{Experiment}

\subsection{Dataset and Evaluation Metrics} 

\textbf{Dataset} Our experiments are conducted on the large-scale Interhand2.6M dataset \cite{moon2020interhand2}, which consists of about 1.3M training images and 0.8M test images. We use all single-hand (SH) and interacting-hand (IH) images in the training set for training. All images are cropped and resized to $256\times256$ based on the bounding boxes provided by the dataset. We adopt the widely-used mean per-joint position error (MPJPE) and mean per-vertex position error (MPVPE) as the evaluation metrics. 

\subsection{Implementation Details} Our networks are trained with 4 GeForce RTX 4090 graphics cards. We adopt the Adam optimizer \cite{kingma2014adam} with a batch size of 120 and 30 training epochs. Each epoch takes about five hours. The initial learning rate is $10^{-3}$ and is scaled by 0.5 after each epoch until it reaches $10^{-6}$. Detailed settings of hyperparameters are given in the supplemental material on our project page.

We adopt several data augmentation methods to improve the generalization ability of the proposed network following \cite{li2022interacting}, including random image shifting in $[-10, 10]$ pixels, random rotations in $[-90\degree, 90\degree]$, random resizing with scaling factor in $[0.9, 1.1]$, horizontal flipping, and adding Gaussian noise $\sim \mathcal{N} (0, 0.3)$ to images.

\subsection{Comparison with State-of-the-arts}
\begin{table}[!t]
    \caption{MPJPE (mm $\downarrow$) on Interhand2.6M.}
    \label{tab:Comparison}
    \centering
    \begin{tabular}{|c|c|c|c|}
    \hline
    Method & SH & IH & All \\ \hline
    \citealt{moon2020interhand2} & 12.16 & 16.02 & 14.22 \\
    \citealt{zhang2021interacting} & - & 13.48 & - \\
    \citealt{hampali2022keypoint} & 10.99 & 14.34 & 12.78 \\
    \citealt{meng20223d} & 8.51 & 13.12 & 10.97 \\
    \citealt{li2022interacting} & - & 10.13 & - \\
    \citealt{yu2023overcoming} & - & 9.68 & - \\
    \citealt{lee2023im2hands} & - & 9.68 & -  \\
    \citealt{jiang2023a2j} & 8.10 & 10.96 & 9.63 \\
    \hline
    Ours-MLP & 7.84 & 10.53 & 8.78 \\ 
    Our-GraphAttn & 7.23 & 9.55 & 8.40   \\  \hline
    %/data1/lihanhui/intaghand/CodeBK/0507_still_have_mid/full_ugt_1.0_05_08_before_part
    \end{tabular}
\end{table}

\begin{figure*}[!t]
  \centering
  \includegraphics[width=\textwidth]{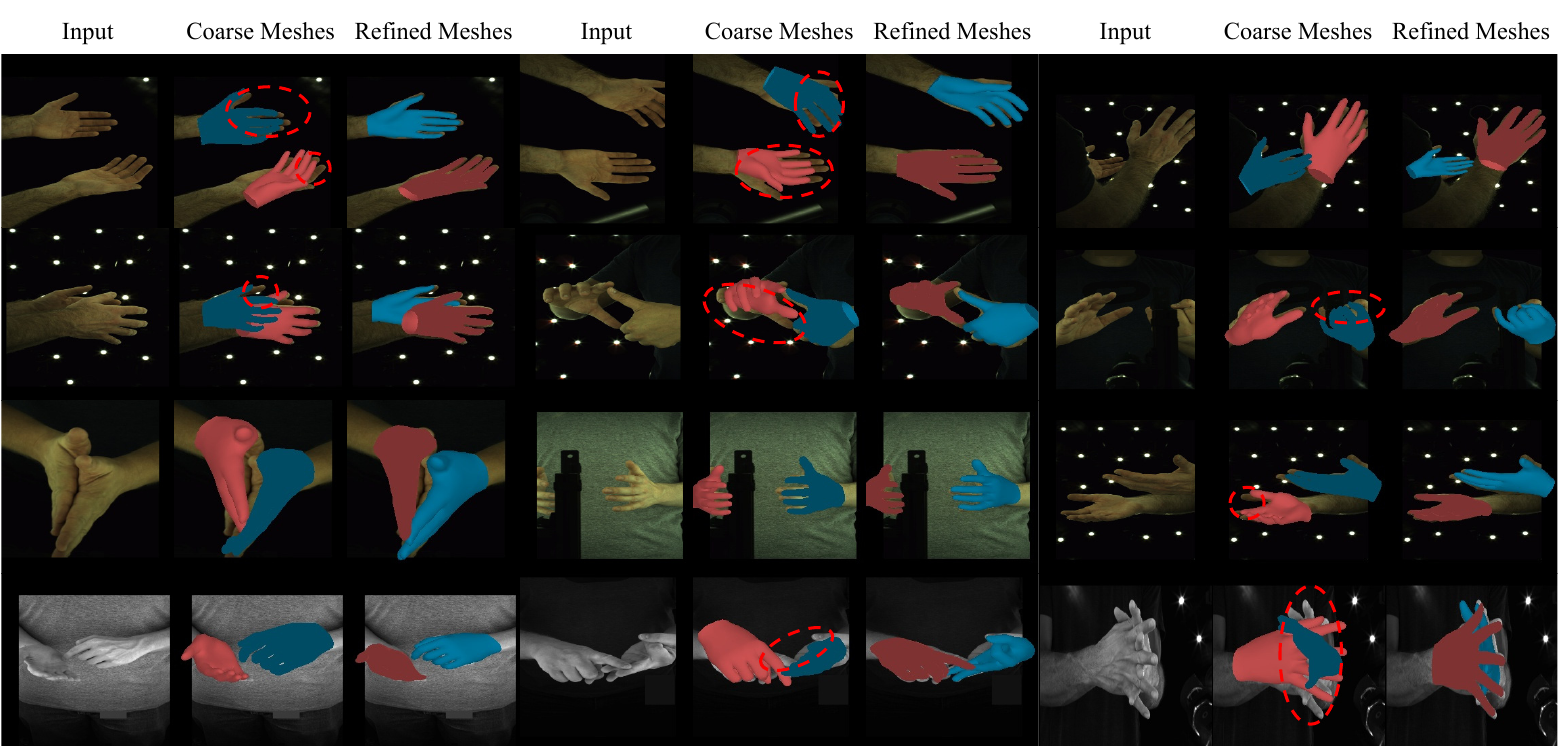}
  \caption{Visual comparison between the coarse meshes obtained by fitting the MANO model and those refined by the proposed method. Best viewed on screen.}
  \label{fig:qualablation}
\end{figure*}

Table \ref{tab:Comparison} reports the MPJPE performance of the proposed method against cutting-edge methods. As we have emphasized above, our dual noise estimation process does not have any other constraint besides the MANO topology, and hence it can be combined with different methods. To validate this, we implement two network variants, i.e., the one with the proposed multi-view MLPs (denoted as Ours-MLP) and the other one with graph attentions (\citealt{li2022interacting}, denoted as Ours-GraphAttn). From Table \ref{tab:Comparison} we can see that these two variants achieve state-of-the-art performance by reducing the MPJPE on the Interhand2.6M dataset from $9.63$ to $8.78$ (Ours-MLP) and $8.40$ (Ours-GraphAttn). This indicates that the proposed dual noise estimation is a considerable strategy for current hand mesh recovery methods.

Figure \ref{fig:qualcomp} demonstrates the visual examples of interacting hand meshes reconstructed by our method (Ours-MLP) and the state-of-the-art IntagHand method \cite{li2022interacting}. It is reasonable that IntagHand can generate well-aligned results because it is a non-parametric model. However, without the geometric prior of hands, it still exhibits artifacts near the fingertips or inaccurate 2D projections (e.g., the middle example in Figure \ref{fig:qualcomp}). On the contrary, the proposed method adopts the coarse-to-fine paradigm, in which the coarse but relatively reliable hand meshes are exploited and refined progressively. Hence the reconstruction results of the proposed method are more stable and accurate.

\subsection{Ablation Study}

To provide a comprehensive analysis of the proposed method, we conduct several ablation experiments in this section. Considering that network training on the whole training set of Interhand2.6M is computationally expensive, we only use $10\%$ of training data in our ablation studies, and the whole test set is used for evaluation. Other experimental settings remain unchanged in our ablation studies.

% \begin{table}[t]
%     \caption{Ablation study on the proposed modules. MPVPE2 and MPVPE3 denote the MPVPE metric calculated on 2D and 3D predictions.}
%     \centering
%     \resizebox{0.495\textwidth}{!}{
%     \begin{tabular}{|c|c|c|c|c|}
%     \hline
%     Method & MPVPE3-SH & MPVPE3-IH & MPVPE2-SH & MPVPE2-IH \\ 
%     \hline
%     Baseline & 11.45 & 14.51 & 13.92 & 12.18 \\ 
%     $ + \varepsilon_{2d}$ & - & - & 12.93 & 11.54 \\ 
%     $ + \varepsilon_{3d}$ & 10.02 & 12.89 & 10.20 & 9.96 \\
%     Full DNE & 9.89 & 12.68 & 9.64 & 9.06 \\ \hline
%     \end{tabular}}
%     \label{tab:ablation}
% \end{table}

\textbf{Effects of 3D noise estimation}. We first evaluate the effectiveness of 3D noise estimation. This experiment considers three variants of the proposed method: the parametric fitting baseline, the baseline augmented with 3D noise estimation (denoted as $ + \varepsilon_{3d}$), and the full DNE module. The experimental results are reported in Table \ref{tab:ablation_3d}. $ + \varepsilon_{3d}$ brings notable performance gains to the baseline, as it reduces the MPVPE of the baseline on the sing-hand subset/interacting-hand subset/whole test set from 11.45/14.51/12.53 to 10.02/12.89/11.02. This suggests that the proposed multi-view MLP based 3D noise estimation module is effective. The full DNE module further improves the performance of the baseline and outperforms the variant with 3D noise estimation only on all test subsets. We owe this to the proposed 2D noise estimation and camera correction methods, as they help to obtain features that are more aligned with images. 

\begin{figure*}[!t]
  \centering
  \includegraphics[width=\textwidth]{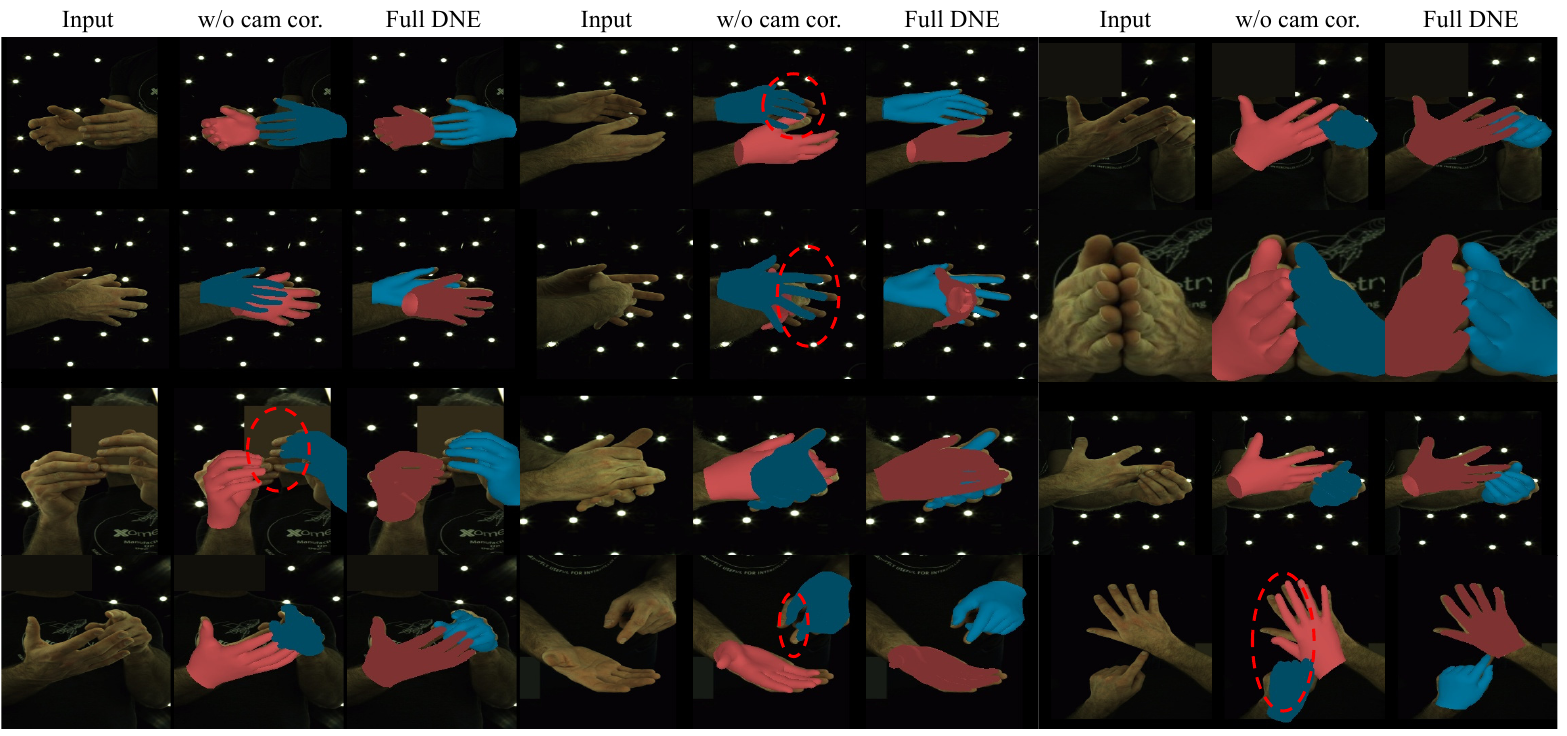}
  \caption{Visual comparison between the DNE module with and without the camera correction. Best viewed on screen.}
  \label{fig:qualcam}
\end{figure*}

\begin{table}[t]
    \caption{Ablation study on 3D noise estimation.}
    \centering
    \begin{tabular}{|c|c|c|c|}
    \hline
    Method & MPVPE-SH & MPVPE-IH & MPVPE-All \\ 
    \hline
    Baseline & 11.45 & 14.51 & 12.53 \\ 
    $ + \varepsilon_{3d}$ & 10.02 & 12.89 & 11.02  \\
    Full DNE & 9.89 & 12.68 & 10.87 \\ \hline
    \end{tabular}
    \label{tab:ablation_3d}
\end{table}

\textbf{Effects on 2D noise estimation} We are also interested in the effects of each proposed component on 2D predictions, as generating well-aligned results is one of the major goals of this paper. Besides the three variants used in the previous experiment, we consider another variant that estimates 2D noise only (denoted as $ + \varepsilon_{2d}$). In this ablation study, we use 2D MPVPE as the metric, and the results of these four variants are summarized in Table \ref{tab:ablation_2d}. We observe that both $ + \varepsilon_{2d}$ and $ + \varepsilon_{3d}$ outperform the baseline and the performance margin of the latter is more obvious. This is reasonable, as the conceptual field of a vertex in the 3D noise estimation process (max pooling on three-view feature maps) is larger than that in the 2D case (only features interpolated with the projected and the regressed coordinates). Consequently, the 3D noise estimation module can leverage more information for refinement. Utilizing 2D and 3D noise estimation jointly achieves the best performance. The full DGE module reduces the MPVPE of the baseline by more than $30\%$/$25\%$/$27\%$ on the three test sets, respectively. These results are sufficient to validate that the proposed method boosts the performance of the baseline in image-plane alignment successfully.

\begin{table}[t]
    \caption{Ablation study of the proposed modules on 2D predictions. The MPVPE is calculated with 2D coordinates.}
    \centering
    \begin{tabular}{|c|c|c|c|}
    \hline
    Method & MPVPE-SH & MPVPE-IH & MPVPE-All \\ 
    \hline
    Baseline & 13.92 & 12.18 & 13.31 \\ 
    $ + \varepsilon_{2d}$ & 12.93 & 11.54 & 12.45 \\ 
    $ + \varepsilon_{3d}$  & 10.20 & 9.96 & 10.12 \\
    Full DNE & 9.64 & 9.06 & 9.44 \\ \hline
    \end{tabular}
    \label{tab:ablation_2d}
\end{table}

% \begin{table}[t]
%     \caption{Ablation study on the number of DNE modules.}
%     \centering
%     \resizebox{0.47\textwidth}{!}{
%     \begin{tabular}{|c|c|c|c|c|}
%     \hline
%     Method & MPVPE-SH & MPVPE-IH & MPJPE-SH & MPJPE-IH \\ 
%     \hline
%     Baseline & 11.45 & 14.51 & 11.15 & 14.10 \\ 
%     M = 1 & 10.66 & 13.52 & 10.33 & 13.16 \\
%     M = 3 & 9.89 & 12.68 & 9.69 & 12.41 \\ \hline
%     \end{tabular}}
%     \label{tab:ablation_number2}
% \end{table}

\textbf{Effects on the number of DNE modules}. At last, as we have mentioned above, the mesh refinement process can be conducted progressively via multiple DNE modules. To verify this, we compare the performance of using a single DNE module ($M=1$) and that of using three DNE modules ($M=3$). The experimental results are reported in Table \ref{tab:ablation_number}. These results validate that higher performance gains can be obtained with more DNE modules. 

\begin{table}[t]
    \caption{Ablation study on the number of DNE modules.}
    \centering
    \begin{tabular}{|c|c|c|c|}
    \hline
    Metric & Baseline & M = 1 & M = 3 \\ 
    \hline
    MPVPE-SH & 11.45 & 10.66 & 9.89 \\
    MPVPE-IH & 14.15 & 13.52 & 12.68 \\
    MPVPE-All & 12.53 & 11.67 & 10.87\\
    MPJPE-SH & 11.15 & 10.33 & 9.69 \\
    MPJPE-IH & 14.10 & 13.16 & 12.41 \\
    MPJPE-All & 12.19 & 11.33 & 10.64 \\
    \hline
    \end{tabular}
    \label{tab:ablation_number}
\end{table}

\textbf{Visual comparison of mesh refinement}. The visual comparison between the meshes before and after refinements is shown in Figure \ref{fig:qualablation}. We can see that the meshes refined by the proposed method are more accurate. This again validates that leveraging the advantages of parametric models and non-parametric models is considerable. 

\textbf{Visual comparison of camera correction}. We also compare the reconstruction results of the proposed with and without the camera correction in Figure \ref{fig:qualcam}. From this figure, we can see that leveraging camera correction helps to generate better results.

\section{Conclusion}
In this paper, we propose a novel method leveraging dual noise estimation to recover 3D hand meshes from single-view images. Our method models the noise of mesh vertices and their projections on the image plane in a unified probabilistic model. We implement the proposed framework via an end-to-end trainable network with two effective estimation branches. Furthermore, our framework can also refine the intrinsic camera parameters efficiently via ridge regression. Consequently, our method can generate hand meshes that are well-aligned with images. Our experiments and ablation studies on the Interhand2.6M dataset demonstrate the effectiveness of our method.

Our current method is not designed especially for single-hand or interacting-hand images. In the future, we plan to incorporate a cross-hand noise model to further enhance the proposed method. We will also consider other association strategies for vertices and their 2D coordinates, such as differential neural rendering.

\section{Acknowledgments} 
This work was supported in part by National Key R\&D Program of China under Grant No. 2020AAA0109700,  Guangdong Outstanding Youth Fund (Grant No. 2021B1515020061), National Natural Science Foundation of China (NSFC) under Grant No. 61976233, No. 92270122, No. 62372482 and No. 61936002, Mobility Grant Award under Grant No. M-0461, Shenzhen Science and Technology Program (Grant No. RCYX20200714114642083), Shenzhen Science and Technology Program (Grant No. GJHZ20220913142600001), Nansha Key R\&D Program under Grant No.2022ZD014 and Sun Yat-sen University under Grant No. 22lgqb38 and 76160-12220011.

\bibliography{aaai24}

\end{document}